\documentclass[11pt]{article}
\usepackage[margin=1in]{geometry}
\usepackage{graphicx} % Required for inserting images
\usepackage{multirow}
\usepackage{caption}
\usepackage{subcaption}
\usepackage{colortbl}
\usepackage{amsmath,amsfonts,amssymb}
\usepackage{url}
\usepackage{hyperref}
\usepackage{booktabs}

\title{Multiple-Choice Question Generation Using Large Language Models: Methodology and Educator Insights}

\author{Giorgio Biancini\thanks{gio.biancini@stud.uniroma3.it} \and 
        Alessio Ferrato\thanks{alessio.ferrato@uniroma3.it} \and 
        Carla Limongelli\thanks{limongel@dia.uniroma3.it}}

\begin{document}

\maketitle

\begin{abstract}
Integrating Artificial Intelligence (AI) in educational settings has brought new learning approaches, transforming the practices of both students and educators. Among the various technologies driving this transformation, Large Language Models (LLMs) have emerged as powerful tools for creating educational materials and question answering, but there are still space for new applications. Educators commonly use Multiple-Choice Questions (MCQs) to assess student knowledge, but manually generating these questions is resource-intensive and requires significant time and cognitive effort. In our opinion, LLMs offer a promising solution to these challenges. 
This paper presents a novel comparative analysis of three widely known LLMs - Llama 2, Mistral, and GPT-3.5 - to explore their potential for creating informative and challenging MCQs. In our approach, we do not rely on the knowledge of the LLM, but we inject the knowledge into the prompt to contrast the hallucinations, giving the educators control over the test's source text, too. Our experiment involving 21 educators shows that GPT-3.5 generates the most effective MCQs across several known metrics. Additionally, it shows that there is still some reluctance to adopt AI in the educational field.
This study sheds light on the potential of LLMs to generate MCQs and improve the educational experience, providing valuable insights for the future.\\
\\
\textit{This paper was accepted for publication in the Adjunct Proceedings of the 32nd ACM Conference on User Modeling, Adaptation and Personalization (UMAP Adjunct '24), July 1--4, 2024, Cagliari, Italy.}
\textit{© ACM 2024. This is the author's version of the work. It is posted here for your personal use. Not for redistribution. The definitive Version of Record was published in Adjunct Proceedings of the 32nd ACM Conference on User Modeling, Adaptation and Personalization (UMAP Adjunct '24), http://dx.doi.org/10.1145/3631700.3665233.}
\end{abstract}

\section{Introduction}

Incorporating Artificial Intelligence (AI) into education can potentially transform the teaching and learning landscape. However, discussions surrounding AI in this context are frequently characterized by a blend of excitement and doubt. A significant issue revolves around the readiness of educators to employ these technologies in a meaningful manner. The unwillingness to embrace new tools stems from educators' apprehension about changing established teaching methodologies due to their unfamiliarity with new technologies. From our perspective, this gap presents a distinct opportunity for AI to offer numerous advantages to educators. For example, AI can assist teachers to~\cite{whalen2023chatgpt}:
\begin{itemize}
\item Streamlining the drafting of course materials, including syllabuses, lesson plans, and learning objectives.
\item Enhancing teaching strategies through suggestions for inclusive activities and diverse reading lists.
\item Supporting personalized student learning with individualized tutoring and simplified explanations of complex topics.
\item Assisting in creating practice test questions and quizzes, thus easing the assessment process.
\end{itemize}
One significant application of AI concerns the last point. In particular, Large Language Models (LLMs) can automate the creation of Multiple Choice Questions (MCQs), which traditionally demand considerable time and effort from educators.

By focusing on using LLMs to generate MCQs, this study diverges from the existing literature, which predominantly explores LLMs for question-answering. Our research specifically investigates the effectiveness of LLMs, such as Llama 2, Mistral, and GPT-3.5, in generating MCQs. Through a comparative analysis conducted via a user study, we aim to explore the potential applications of LLMs in this specific task.

\section{Background} 

Questions are crucial for evaluating a learner's knowledge and understanding. Among various testing formats, MCQs stand out for their efficiency in assessment, uniformity in scoring, and suitability for electronic evaluation~\cite{Rao:2020survey}. MCQs help assess a student's recall and understanding of specific information or knowledge in the text~\cite{araki2016generating}.
An MCQ consists of three basic components:
\begin{itemize}    
    \item \textit{Stem}: The part of the question that presents the context or situation on which the question is based.
    \item \textit{Key}: The correct answer of the question. 
    \item \textit{Distractors}: The incorrect options proposed as possible answers are intended to confuse or distract students.
\end{itemize}
The manual MCQ preparation involves understanding the input text, identifying informative sentences with potential questions, selecting the keyword or phrase as the answer, forming a question around it, and choosing distractors from the text or related context. Crafting MCQs manually is labor-intensive and costly~\cite{gamage2019}; these aspects encourage research into automating their generation.
The work of Rao et al.~\cite{Rao:2020survey} delves into this process, examining key stages employed to generate MCQs. For the authors, the automatic generation approach mirrors manual methods. It is structured into stages: (1) \textit{pre-processing} to prepare the text, (2) \textit{sentence selection} to identify suitable content, (3) \textit{key selection} for the answer, (4) \textit{question generation}, (5) \textit{distractor selection} for wrong answers, and (6) \textit{post-processing} to finalize the questions. This streamlined workflow is commonly adopted across various systems, albeit with minor variations. This field has attracted significant interest, with the foundational work on automatic MCQ generation laid by Coniam David in 1997, leveraging word tagging and frequencies to create various types of MCQs~\cite{coniam1997preliminary}. 

Subsequent advancements have witnessed the proliferation of MCQ generation systems across diverse languages and disciplines. Building upon this foundation, Miktov et al.~\cite{mitkov2006computer} directed their efforts towards generating questions within the medical domain, employing techniques such as shallow parsing, term extraction, sentence transformation, and computation of semantic distance. They utilized corpora and ontologies (i.e., WordNet) for different process stages. Majumder et al.~\cite{majumder2015system} primarily focused on sentence selection based on topic word and parse structure similarity within the realm of sports. Maheen et al.~\cite{maheen2022automatic} endeavored to generate computer science questions utilizing classical machine learning and text embedding techniques with BERT. Their system underwent validation by ten domain experts across three metrics (i.e., sentence selection, key generation, and distractor selection). Notably, both approaches exhibit several intermediary steps, potentially restraining their application across domains beyond those for which they were originally designed. 

The quality of MCQs is heavily contingent on the quality of distractors. Insufficiently challenging distractors may lead examiners to identify the correct answer easily. In this context, the authors of~\cite{kumar2023novel} and~\cite{lakshmi:2023multiple} propose systems utilizing BERT for text processing and WordNet for generating distractors. A recent contribution on language models is presented in~\cite{bulathwela2023scalable} where authors demonstrated how pre-trained language models can be customized for generating questions in the educational context outlining also the interest in MCQ generation. 

The primary challenges in automatic MCQ generation lie in identifying and adapting the most effective systems capable of executing the intermediary steps necessary for generating desired questions, as emphasized by Rao~\cite{Rao:2020survey}. However, we believe that LLMs are capable of autonomously perform all the necessary steps for MCQ generation even without additional training.
Despite their potential, the existing literature on LLMs and MCQs has predominantly focused on evaluating these models for question-answering tasks (e.g.,~\cite{robinson2022leveraging}). Only recently the author of~\cite{meissner2024} have proposed a system prototype for the automated generation of self-assessment quizzes validating it with 6 participants.
Extending their work, we want to  evaluate LLMs for MCQ generation employing a versatile prompt applicable under various conditions and domains since it does not rely on the knowledge of the LLM. We conduct an experiment with 21 domain experts, including high school and university professors, utilizing several established metrics to assess the quality of generated MCQs from~\cite{Rao:2020survey}.

\section{Prompt for MCQ generation}
\label{sec:prompt}

According to existing research, constructing queries for LLMs, known as prompt engineering, poses significant challenges~\cite{zamfirescu2023johnny}. Achieving desired responses from the model requires multiple iterations, making it complex. We dedicated significant effort to crafting a prompt that adheres to a strict policy for eliciting responses. The definitive query used across all LLMs in our research is as follows:

{\footnotesize
\begin{verbatim}
You are an assistant that doesn't make mistakes.
If a reference model is presented to you, you follow it perfectly without making errors.

Create a high school - level quiz based on the provided text.
You must strictly adhere to the following format without any errors:
> [ Insert the question ]
a ) [ Option A ]
b ) [ Option B ]
c ) [ Option C ]
d ) [ Option D ]
* Correct Answer: [Insert the letter corresponding to the correct answer for example : 'a)']
* Source: [Write the exact line or passage from the provided text where the information for this question can be found.]

Please note that you are allowed to modify only the parts within brackets ([...]) in the format provided.
Ensure that all four options are distinct.
When mentioning a date, please make sure to specify the year.
The text is : { text }
\end{verbatim}
}
We insert a paragraph from the text serving as the knowledge base for the model into the "text" parameter. This query generates questions proportional to the paragraph length, implying that longer paragraphs may yield more questions.

\section{Experiment}

\subsection{Setting}

\subsubsection{Metrics} 
\label{sec:metrics}
The experimentation aims to test which of the three LLMs better meets the following evaluation criteria described in~\cite{Rao:2020survey}:
\begin{enumerate}
    \item compliance of the question with the source text provided (compliance or sentence selection),
    \item clarity of formulation (clarity),
    \item selection of the proposed distractors (distractors selection),
    \item the soundness of the correct answer with respect to the question (coherence of key selection),
    \item utility for learning (learning utility).
\end{enumerate}

\subsubsection{Sample selection}
We distributed the form to a diverse group of high school and university teachers with varying levels of experience and expertise across disciplines. The sample included educators skilled in designing educational materials and evaluating student progress, regardless of their area of study.

\subsubsection{Questionnaire generation}

The topic considered for the experiment regards the Second World War\footnote{We used the following knowledge source \url{https://en.wikipedia.org/wiki/Causes_of_World_War_II} (last access: \today)}, an acknowledged piece of history.
We employed the prompt presented in~\ref{sec:prompt} for each paragraph of the text. As we did not impose a set number of questions, the number of multiple-choice questions varied from model to model and paragraph to paragraph. To create the test, we randomly selected seven MCQs for each LLM. All generations were conducted using Google Colab's free tier with Llama 2\footnote{\url{https://huggingface.co/meta-llama/Llama-2-7b-chat-hf} (last access: \today)} and Mistral\footnote{\url{https://huggingface.co/mistralai/Mistral-7B-Instruct-v0.1} (last access: \today)} from HuggingFace. Both models used a temperature of 0.5, a maximum of 2048 new tokens, and a top k value of 30. Other hyperparameters were set to default.
For generations using GPT-3.5, we used ChatGPT\footnote{\url{https://chat.openai.com/} (last access: \today)} and provided a new prompt for every paragraph while omitting the chat history from the account settings.

\subsubsection{Questionnaire execution}
The online form was created using \textit{Google Forms} and was about 30 minutes long. The form contained 21 (i.e., seven for each LLM) distinct generated MCQs. For each one there were the text sourrce (i.e., the piece of text that the LLM used to construct the question), the question with four the possible answers, and the correct answer. Respondents were asked to rate the five criteria established in \ref{sec:metrics} on a five-point Likert scale, ranging from "strongly disagree" to "strongly agree." We also asked participants about their willingness to use LLMs to generate MCQs. This comprehensive approach allowed for a detailed assessment of the LLMs' performance and provided valuable insights into the perceptions of participants. We did not collect any demographic information about the teachers.

\subsection{Results}
\begin{table}[t]
\centering
\caption{Descriptive statistics and Friedman test results. The extended description of categories is reported in \ref{sec:metrics}}
\label{tab:anova_results}
\begin{tabular}{@{}lllllll@{}}
\toprule
Metric & LLM & Mean & Std & Median & F-statistic & p-value \\ \midrule
\multirow{3}{*}{Clarity} & GPT-3.5 & \textbf{1.496599} & 0.822410 & 2 & \multirow{3}{*}{23.490066} & \multirow{3}{*}{0.000008*} \\
 & Llama & 1.292517 & 0.959530 & 2 &  &  \\
 & Mistral & 0.986395 & 1.193512 & 1 &  &  \\ \midrule
\multirow{3}{*}{Coherence} & GPT-3.5 & \textbf{1.013605 }& 1.394639 & 2 & \multirow{3}{*}{18.369620} & \multirow{3}{*}{0.000103*} \\
 & Llama & 0.646259 & 1.456223 & 1 &  &  \\
 & Mistral & 0.421769 & 1.456799 & 1 &  &  \\ \midrule
\multirow{3}{*}{Compliance} & GPT-3.5 & \textbf{1.346939} & 1.083203 & 2 & \multirow{3}{*}{60.903382} & \multirow{3}{*}{0.000000*} \\
 & Llama & 0.360544 & 1.608818 & 1 &  &  \\
 & Mistral & -0.006803 & 1.563700 & 0 &  &  \\ \midrule
\multirow{3}{*}{Distractor selection} & GPT-3.5 & \textbf{1.006803} & 1.082214 & 1 & \multirow{3}{*}{14.116279} & \multirow{3}{*}{0.000860*} \\
 & Llama & 0.748299 & 1.198421 & 1 &  &  \\
 & Mistral & 0.537415 & 1.195151 & 1 &  &  \\ \midrule
\multirow{3}{*}{Learning utility} & GPT-3.5 & \textbf{0.625850} & 1.277899 & 1 & \multirow{3}{*}{1.749373} & \multirow{3}{*}{0.416993} \\
 & Llama & 0.605442 & 1.213835 & 1 &  &  \\
 & Mistral & 0.489796 & 1.212759 & 1 &  & \\ \bottomrule
 \textit{* p-value $<$ 0.005}

\end{tabular}%
\end{table}

We utilized the non-parametric Friedman test to assess whether the responses provided by 21 educators varied significantly across the three models analyzed. Our assumptions were as follows: 
\begin{itemize}
   \item  Null hypothesis:  LLMs are equally good in generating MCQs on a given metric;
    \item Alternative hypothesis: There is at least one LLM better than the others in generating MCQs.
\end{itemize}

\begin{figure}[!h]
\centering
\begin{subfigure}{.45\textwidth}
  \centering
  \includegraphics[width=0.8\linewidth]{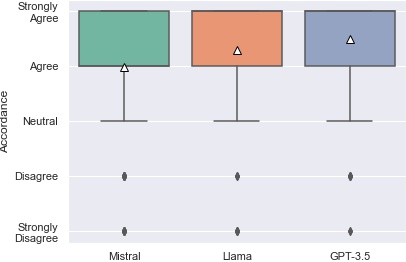}
  \caption{Clarity}
  \label{fig:sfig1}
\end{subfigure}%
\hfill
\begin{subfigure}{.45\textwidth}
  \centering
  \includegraphics[width=0.8\linewidth]{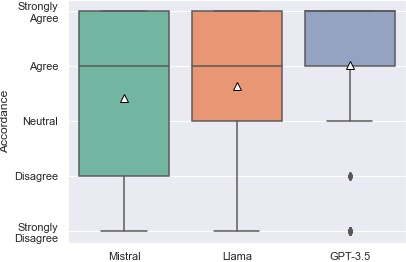}
  \caption{Coherence}
  \label{fig:sfig2}
\end{subfigure}

\vspace{0.5cm}

\begin{subfigure}{.45\textwidth}
  \centering
  \includegraphics[width=0.8\linewidth]{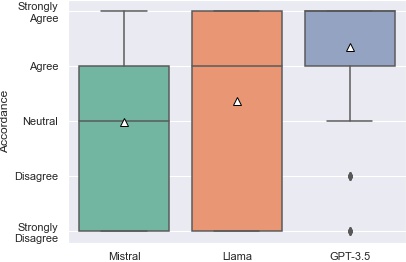}
  \caption{Compliance}
  \label{fig:sfig3}
\end{subfigure}%
\hfill
\begin{subfigure}{.45\textwidth}
  \centering
  \includegraphics[width=0.8\linewidth]{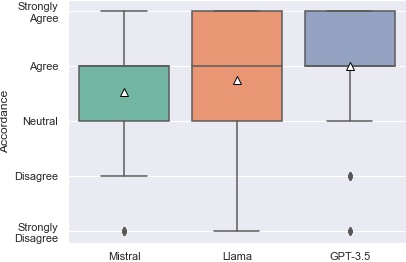}
  \caption{Distractor selection}
  \label{fig:sfig4}
\end{subfigure}

\vspace{0.5cm}

\hfill
\begin{subfigure}{.99\textwidth}
  \centering
  \includegraphics[width=0.4\linewidth]{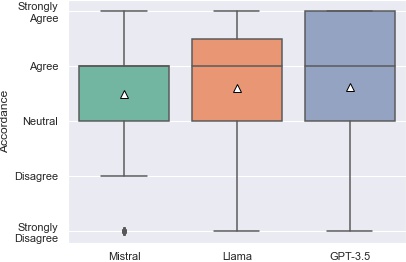}
  \caption{Learning utility}
  \label{fig:sfig5}
\end{subfigure}
\hfill

 \caption{Distribution of the responses for each LLM and metric. The small white triangles indicate the mean value for the box.}
\label{fig:grafici}
\end{figure}

\begin{figure}[ht]
\centering
\includegraphics[width=0.38\linewidth]{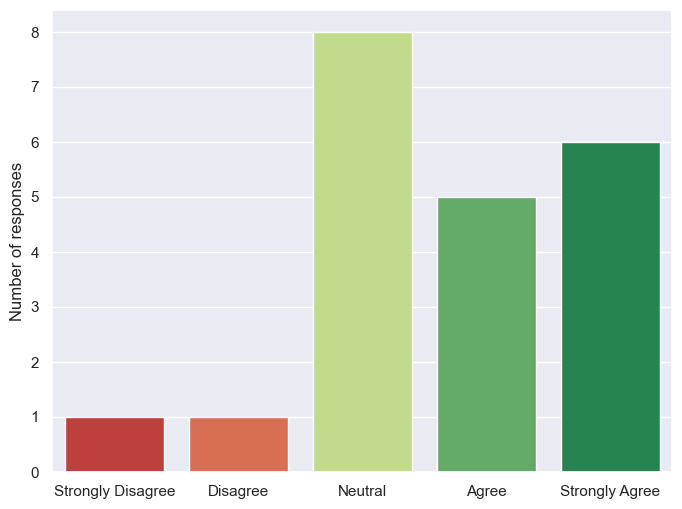}
    \caption{Distribution of the responses regarding the possible adoption of automatic MCQ generators by educators.}
    \label{fig:bars}
\end{figure}

We have decided to use this particular test since the Shapiro-Wilk test rejected the normality distribution of the data, and each participants voted for the MCQs of all three LLMs. We opted for this test since the data did not follow a normal distribution, as indicated by the Shapiro-Wilk test, and the participants on the MCQs generated evaluated all three LLMs. 

The results, presented in Table \ref{tab:anova_results}, showed significant differences in the evaluations of the models, except for "Learning utility," where the difference was not statistically significant, which, in our opinion, can be influenced by various external and internal factors specific to each individual. It is crucial to acknowledge these inherent challenges when assessing this metric. Despite this, upon closer examination of the data, as illustrated in Figure \ref{fig:grafici}, GPT-3.5 consistently outperformed the other two LLMs in MCQ generation across all metrics, with Llama 2 coming in second and Mistral in third. Furthermore, Llama and Mistral exhibited poorer coherence and compliance performance than GPT. This result could potentially be attributed to suboptimal source text or sentence selection. Notably, both models at times identified the position of the sentence as the source text instead of the actual sentence itself (e.g. "Last sentence of the text" or "Line 5 of the text"). This affected both metrics because it was difficult for participants to assess that the answer reported as correct actually was correct by not having the text from which the question was created.
It is worth mentioning that GPT did not exhibit such a phenomenon.

In the last survey question, we asked participants about possibly using this or a similar system. Based on the responses, as shown in Figure \ref{fig:bars}, the results were varied. While only two out of 21 participants showed disinterest in using the system, eleven participants were in favor of adopting it, and eight remained neutral. These mixed findings shed light on opportunities and potential problems with this system. It is essential to conduct additional research on the factors that encourage or discourage the acceptance of such systems in the future. This will aid in comprehending the motivations behind people's decisions and in building systems that cater to their needs.

\section{Conclusions, limitations and future work}

This paper introduces an innovative methodology for generating MCQs utilizing LLMs. This method diverges from traditional approaches that use LLMs for question answering. Also, with our query, we do not rely exclusively on the intrinsic knowledge base of the LLM. Instead, it employs the LLM as a tool that can execute autonomously the stages involved in automatic MCQs creation. The efficacy of the generated MCQs was evaluated through an experiment conducted with educators. The findings of our investigation indicate that GPT-3.5 is particularly proficient in generating MCQs.

However, this study acknowledges certain limitations. Notably, a comparative analysis between the MCQs generated by LLMs and those publicly available or human-generated was not conducted. Additionally, the scope of our experimentation was limited to a small sample size (i.e., seven questions per LLM and 21 participants), primarily attributed to the extensive duration required for form completion and the requirement of domain experts. The absence of demographic data about the participants also presents a limitation, as such information would have been instrumental in elucidating the reasons behind certain participants' resistance to adopting automated MCQ generators. Another bias concerns the topic used for MCQ generation since there are several domains in which the LLMs have troubles.

While these technologies offer considerable benefits, such as process optimization and increased productivity, they are not without inherent risks, including the presumption of infallibility and potential privacy issues. Initiatives like UNESCO's "Guidance for Generative AI in Education and Research"\footnote{\url{https://www.unesco.org/en/articles/guidance-generative-ai-education-and-research} (last access: \today)} underscore the necessity for establishing regulatory frameworks to ensure the ethical use of AI in educational settings, thereby mitigating potential misuse and ethical concerns. AI's transformative potential can be supported by proactive measures and stakeholder collaboration, adhering to principles of accountability and ethical integrity in education.

It is crucial to perceive AI as a supplementary instrument rather than a substitute for human expertise~\cite{grassini2023shaping}. Hence, future research should aim to develop autonomous frameworks that enable the seamless incorporation of LLMs into educational environments for MCQ generation. Additionally, the potential application of Bloom's Taxonomy~\cite{anderson2001taxonomy} could lead to a differentiation of the MCQ's learning objectives~\cite{elkins2023useful} that could also enhance the student's learning experience. 

Furthermore, following what has been done in recommender systems~\cite{Bologna2013,ferrato23,Sansonetti19}, implementing personalization based on users' personalities or emotional states could improve the efficacy of AI-driven educational frameworks~\cite{conati2021toward}. Automatic MCQ generation goes beyond traditional educational settings and can be used in various contexts. For example, after cultural visits~\cite{DeAngelis17}, such systems could ask users about the main points of interest that caught their attention~\cite{Ferrato22}, thus creating more meaningful experiences.

We recognize that LLMs could hallucinate~\cite{zhang2023siren}. Addressing this issue should be a priority for future developments. While methods like Knowledge Injection~\cite{martino2023knowledge} have been proposed, and we have partially implemented it using custom text (i.e., the injected paragraph) as a knowledge base to mitigate this problem, more efforts should be dedicated to this approach. Another technique used to address this issue is the application of Retrieval-Augmented Generation~\cite{lewis2020retrieval} to LLMs, allowing them to utilize trustworthy knowledge bases that are not used to train the model. Although this technique has been used in question answering~\cite{sun24}, it can also be applied in open-ended questions and MCQ generation.

In the future, our goal is to overcome the limitations of this study and advance with creating an open-source tool for supervised MCQ generation that educators and researchers can utilize. Implementing similar systems to the one demonstrated in this work can significantly improve educational practices as long as they are utilized responsibly.

\bibliographystyle{alpha}
\bibliography{ref}

\appendix
\section{Questions}
Here, we show the randomly selected questions used in the questionnaire.

\begin{table}[ht]
\tiny
\caption{MCQs generated by GPT-3.5}
\label{tab:appendix1}
\resizebox{\textwidth}{!}{%
\begin{tabular}{p{6cm}p{2cm}ll}
\toprule
Source & Question & Options & Correct \\ \midrule
The British declared war on Japan on the morning of 8 December and specifically identified the attacks on Malaya, Singapore and Hong Kong as the cause but omitted any mention of Pearl Harbor. & What event caused Canada to declare war on Japan? & \begin{tabular}[t]{@{}l@{}}a) The attacks on Pearl Harbor\\ b) The attacks on Malaysia, Singapore and Hong Kong\\ c) The sinking of the Lusitania\\ d) Germani's invasion of Poland.\end{tabular} & b \\  \\
Hitler chose to declare that the Tripartite Pact required Germany to follow Japan's declaration of war although American destroyers escorting convoys and German U-boats had been de facto at war in the Battle of the Atlantic. & What was the purpose of the Tripartite Pact? & \begin{tabular}[t]{@{}l@{}}a) Establish a military alliance between Germany, Italy and Japan\\ b) Establish a military alliance between Germany, Italy and the United States\\ c) Establish a military alliance between Germany, Italy and the Soviet Union\\ d) Establish a military alliance between Germany, Italy and the United Kingdom\end{tabular} & a \\ \\
Hitler chose to declare that the Tripartite Pact required Germany to follow Japan's declaration of war although American destroyers escorting convoys and German U-boats had been de facto at war in the Battle of the Atlantic. & What event brought the United States into the European war? & \begin{tabular}[t]{@{}l@{}}a) The United States declared war on Germany and Italy.\\ b) Germany and Italy declared war on the United States.\\ c) The United States joined the Tripartite Pact\\ d) The United States joined the Axis powers.\end{tabular} & b \\ \\
Japan stepped in as a mediator in the war in May 1941 and allowed its ally to occupy the bordering provinces in Cambodia and Laos. & What happened in May 1941 during the Franco-Thai War? & \begin{tabular}[t]{@{}l@{}}a) Japan intervened as a mediator in the war and allowed its ally to occupy the\\ neighboring provinces of Cambodia and Laos.\\ b) The Soviet Union launched a surprise attack on Germany.\\ c) The United States declared war on Japan.\\ d) The United Kingdom surrendered to Germany.\end{tabular} & a \\ \\
The last larger unit of Polish troops capitulated on October 6, 1939, near Kock, but some units went straight to partisan combat. & What happened to the last largest unit of Polish troops on October 6, 1939? & \begin{tabular}[t]{@{}l@{}}a) They surrendered to the Red Army.\\ b) They were defeated by German forces.\\ c) They went directly to the partisan struggle.\\ d) They were evacuated to neighboring countries.\end{tabular} & a \\ \\
Hitler believed that the Soviet Union could be defeated in a fast and relentless assault that capitalised on the Soviets' ill-prepared state & What was the state of the Soviet Union before Germany attacked it in June 1941? & \begin{tabular}[t]{@{}l@{}}a) Prepared for war \\ b) Ill-prepared for war\\ c) Neutral\\ d) At war with Germany\end{tabular} & b \\ \\
Hitler believed that his success there would bring Britain to the negotiating table, an event which would end the war altogether. & What did Hitler hope to achieve by defeating the Soviet Union? & \begin{tabular}[t]{@{}l@{}}a) Gain more territory\\ b) End the war\\ c) To weaken the Soviet Union\\ d) To gain allies\end{tabular} & b \\\bottomrule
\end{tabular}
}
\end{table}

\begin{table}[ht]
\tiny
\caption{MCQs generated by Llama 2}
\label{tab:appendix2}
\resizebox{\textwidth}{!}{%
\begin{tabular}{p{6cm}p{2.5cm}ll}
\toprule
Source & Question & Options & Correct \\ \midrule
To deter Hitler, Britain and France announced that an invasion would mean war... & What did Britain and France do to deter Hitler from invading Poland? & \begin{tabular}[t]{@{}l@{}}a) They announced that they would declare war on Germany if it invaded Poland.\\ b) They tried to convince the Soviet Union to join deterrence against Hitler.\\ c) They signed defense treaties with Poland and publicly announced their support.\\ d) They tried to negotiate with Hitler and find a peaceful solution.\end{tabular} & a \\  \\
Hitler believed that the Soviet Union could be defeated in a fast and relentless assault that capitalised on the Soviets' ill-prepared state and he hoped that his success there would bring Britain to the negotiating table, an event which would end the war altogether. & What did Hitler hope to achieve by attacking the Soviet Union? & \begin{tabular}[t]{@{}l@{}}a) He wanted to end the war with Britain.\\ b) He wanted to expand Germany's territory.\\ c) It wanted to demonstrate its military strength.\\ d) He wanted to secure a strategic alliance with the Soviet Union.\end{tabular} & d \\ \\
Last line of the text & When did the first combat of the Flying Tigers take place? & \begin{tabular}[t]{@{}l@{}}a) December 7, 1941\\ b) January 23, 1942\\ c) February 15, 1943\\ d) March 1, 1944\end{tabular} & b \\ \\
Line 5 of the text & Why did the United States declared war on Japan? & \begin{tabular}[t]{@{}l@{}}a) Because of the attacks on Malaysia, Singapore, and Hong Kong.\\ b) Because of the attacks on Pearl Harbor\\ c) Because of unprovoked acts of war against the government and people of the United \\States of America\\ d) Because of a combination of reasons\end{tabular} & b \\ \\
Line 3 of the text & What was the U.S. government's response to Japan's invasion of China in 1937? & \begin{tabular}[t]{@{}l@{}}a) The U.S. government imposed a total embargo on all U.S.-Japan trade.\\ b) The U.S. government asked Japan to withdraw its troops from China and Indochina.\\ c) The U.S. government provided military aid to China to help it fight against Japan.\\ d) The U.S. government remained neutral and took no action against Japan's invasion.\end{tabular} & a \\ \\
Last line of the text & Who commanded the Flying Tigers during their first fight after the attack on Pearl Harbor? & \begin{tabular}[t]{@{}l@{}}a) Claire Lee Chennault\\ b) John F.Kennedy\\ c) Douglas MacArthur\\ d) Dwight D. Eisenhower\end{tabular} & a \\ \\
The last larger unit of Polish troops capitulated on October 6, 1939, near Kock... & When did the last largest unit of Polish troops capitulate? & \begin{tabular}[t]{@{}l@{}}a) On September 17, 1939.\\ b) On October 6, 1939.\\ c) In the spring of 1940.\\ d) Never.\end{tabular} & b \\\bottomrule
\end{tabular}%
}
\end{table}

\begin{table}[ht]
\tiny
\caption{MCQs generated by Mistral}
\label{tab:appendix3}
\resizebox{\textwidth}{!}{%
\begin{tabular}{p{4.5cm}p{2cm}ll}
\toprule
Source & Question & Options & Correct \\ \midrule
Canada declared war on Japan on the evening of 7 December, and a royal proclamation affirmed the declaration the next day. & When did the official Japanese declaration of war on the United States and the British Empire appear in Japanese newspapers? & \begin{tabular}[t]{@{}l@{}}a) December 6, 1941\\ b December 7, 1941\\ c) December 8, 1941\\ d) December 9, 1941\end{tabular} & c \\ \\
Hitler admitted to his generals on 23 May 1939 that his reason for invading Poland was not Danzig: "Danzig is not the issue at stake. It is a matter of extending our living space in the East...". & According to Hitler, what was the main reason for invading Poland? & \begin{tabular}[t]{@{}l@{}}a) Gain control of Danzig\\ b) Expand living space in the east\\ c) Establish a military alliance\\ d) Impose economic sanctions\end{tabular} & b \\ \\
Germany invaded Poland on September 1, 1939, and rejected the British and French demands for it to withdraw, which resulted in both to declare war on September 3, 1939, in accordance with the defence treaties with Poland that they had signed and publicly announced. & When did Germany invade Poland, starting World War II? & \begin{tabular}[t]{@{}l@{}}a) August 1, 1939\\ b) September 1, 1939\\ c) October 1, 1939\\ d) November 1, 1939\end{tabular} & b \\ \\
The text mentions, "In September 1940, the Japanese invaded Vichy French Indochina and occupied Tonkin" & In what year did the Japanese invade French Indochina from Vichy? & \begin{tabular}[t]{@{}l@{}}a) 1939\\ b) 1940\\ c) 1941\\ d) 1942\end{tabular} & b \\ \\
On 7 December 1941, without a declaration of war, the Imperial Japanese Navy attacked Pearl Harbor... & On what date did the Imperial Japanese Navy attack Pearl Harbor without declaring war? & \begin{tabular}[t]{@{}l@{}}a) December 6, 1941\\ b) December 7, 1941\\ c) December 8, 1941\\ d) December 9, 1941\end{tabular} & b \\ \\
Reacting to Japanese pressure on French authorities of French Indochina to stop trade with China, the US began restricting trade with Japan in July 1940. & What was the event that prompted the United States to begin restricting trade with Japan in July 1940? & \begin{tabular}[t]{@{}l@{}}a) Japanese invasion of China\\ b) Japanese pressure on French Indochina\\ c) British trade restrictions\\ d) Dutch oil embargo.\end{tabular} & b \\ \\
On 7 December 1941, without a declaration of war, the Imperial Japanese Navy attacked Pearl Harbor with the aim of destroying the main American battle fleet at anchor. & What event occurred on December 7, 1941 without a declaration of war, aiming to destroy the main U.S. battle fleet at anchor? & \begin{tabular}[t]{@{}l@{}}a) The Battle of Midway\\ b) The Doolittle Raid\\ c) The Attack on Pearl Harbor\\ d) The Battle of Okinawa\end{tabular} & c \\ \hline
\end{tabular}
}
\end{table}

\end{document}